\documentclass{Interspeech2024}
\usepackage{adjustbox}
\interspeechcameraready 

\title{Discrete Multimodal Transformers with a Pretrained Large Language Model for Mixed-Supervision Speech Processing}

\name[affiliation={1}]{Viet Anh}{Trinh}
\name[affiliation={2}]{Rosy}{Southwell}
\name[affiliation={1}]{Yiwen}{Guan}
\name[affiliation={1}]{Xinlu}{He}
\name[affiliation={2}]{Zhiyong}{Wang}
\name[affiliation={1}]{Jacob}{Whitehill}

\address{
  $^1$Worcester Polytechnic Institute\\
  $^2$University of Colorado Boulder}
\email{vtrinh@wpi.edu, roso8920@colorado.edu, yguan2@wpi.edu, xhe4@wpi.edu, zhiyong.wang@colorado.edu,jrwhitehill@wpi.edu}


\begin{document}
\maketitle

\begin{abstract}
Recent work on discrete speech tokenization has paved the way for models that can seamlessly perform multiple tasks across modalities, e.g., speech recognition, text to speech, speech to speech translation. Moreover, large language models (LLMs) pretrained from vast text corpora contain rich linguistic information that can improve accuracy in a variety of tasks. In this paper, we present a decoder-only Discrete Multimodal Language Model (DMLM), which can be flexibly applied to multiple tasks (ASR, T2S, S2TT, etc.) and modalities (text, speech, vision). We explore several critical aspects of discrete multimodal models, including the loss function, weight initialization, mixed training supervision, and codebook. Our results show that DMLM benefits significantly, across multiple tasks and datasets, from a combination of supervised and unsupervised training. Moreover, for ASR, it benefits from initializing DMLM from a pretrained LLM, and from a codebook derived from Whisper activations.
\end{abstract}

\section{Introduction}

The recent introduction of discrete speech tokens \cite{zeghidour2021soundstream,barrault2023seamlessm4t} has facilitated the integration of large language model (LLM) capabilities into speech models. Consequently, these speech models not only possess knowledge from the labeled speech-text pairs on which they were trained, but also inherit the extensive linguistic knowledge from the LLM. Furthermore, discrete tokenization opens the possibility for the seamless integration of speech, text, and vision within a single framework, enabling the model to perform multiple translation tasks from one modality to another (see Related Work section below). However, within this new research direction, crucial questions arise regarding how to manage training  when dealing with multi-modal inputs encompassing speech, text, and/or vision; whether mixed supervision (supervised and/or unsupervised) on multiple tasks can benefit prediction accuracy; and how the discrete codebook should be created.

\begin{figure*}[t]
  \centering
\includegraphics[width=\linewidth]{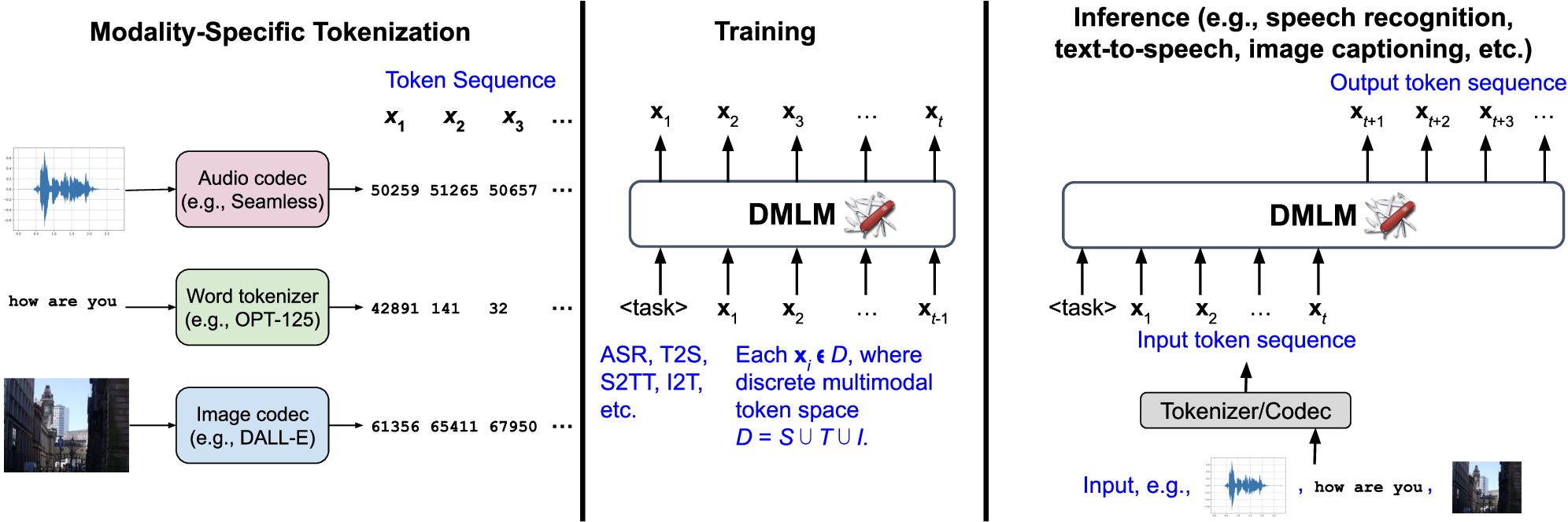}
  \caption{Discrete Multimodal Language Model (DMLM): The input (text, speech, image) is tokenized by a modality-specific tokenizer/codec. The discrete token sequence ${\bf x}_1,{\bf x}_2,\ldots$ is concatenated to a task description and then passed to DMLM for processing. Training is conducted based on next-token prediction. Inference can be performed for a variety of tasks (ASR, S2TT, T2S, I2T, etc.).}
  \label{fig:speech_production}
\end{figure*}

In this paper, we  contribute to the growing field of discrete speech  models  and (1) propose a Discrete Multimodal Language Model (DMLM) (see Figure \ref{fig:speech_production}), which is a unified model that can flexibly translate data from one modality to another to perform multiple tasks for speech processing and more. 
DMLM is one of the first discrete token-based decoder-only models that can translate to and from text, speech, and images. As such, it can benefit from  pretrained large language models, such as OPT \cite{zhang2022opt}, that were optimized on vast text corpora.
We illustrate DMLM on the tasks of automatic speech recognition (ASR), text-to-speech (T2S), speech-to-text translation (S2TT) from 21 different languages, and image captioning (I2T).   In addition, we (2) propose a length-normalized tri-modal loss, a modification of the standard cross-entropy loss for large language models. This adaptation enhances the efficiency of training multi-modal systems, particularly when diverse modalities exhibit significant differences in sequence length.
We (3) conduct extensive experiments to explore
how mixed-supervision training improves performance on ASR, T2S, S2TT and I2T. 
Finally, (4) we introduce a discrete codebook based on Whisper \cite{radford2023robust} and compare with current start-of-the-art codebook \cite{barrault2023seamlessm4t} for the ASR task.

\section{Related Work}
 
\textbf{Discrete Speech Tokens}:
Motivated by advances in text generation via large language models as well as the goal of crafting more robust and space-efficient speech representations, researchers have recently explored ways of discretizing input speech \cite{zeghidour2021soundstream, borsos2023audiolm, barrault2023seamlessm4t}.
One  encoder-decoder method \cite{zeghidour2021soundstream} employs an encoder to  compress speech into embeddings,  and then a decoder to reconstruct  the input speech signal. The embeddings undergo quantization and are expressed as a combination of basis vectors in a codebook. While this method can yield high-quality speech with a specific number of quantizers (e.g., 8 or 12 quantizers), it requires a substantial number of tokens (codebook IDs) to capture just one second of speech -- typically in the range of $8 \times 65$ or $8 \times 75$ discrete tokens per second. An alternative discretization strategy \cite{barrault2023seamlessm4t} involves clustering on a particular layer of a pre-trained speech model, using the clustering ID as the discrete token index.
In particular, \cite{barrault2023seamlessm4t} cluster the activations of a speech model trained through self-supervised learning with contrastive and diversity loss \cite{Babu2021XLSRSC, baevski2020wav2vec}.

\textbf{Speech Processing Based on Pretrained LLMs}:
Advancements in large language models (LLMs) have prompted efforts to leverage pretrained LLMs for speech-related tasks \cite{rubenstein2023audiopalm, barrault2023seamlessm4t}.  For instance, \cite{rubenstein2023audiopalm} utilize PaLM-2 \cite{anil2023palm}, a non-open-source LLM, by fine-tuning it on speech recognition and speech-to-speech translation (S2ST) datasets. However, they employ loss masking on the Automatic Speech Recognition (ASR) task. Loss masking is a special case of the modal-norm loss function that we use (see Section \ref{sec:model}); in particular, it corresponds to setting one or more of the $\lambda$ values to 0. In our work, instead of hard-coding these $\lambda$ to a particular value, we optimize them using hyperparameter search to reach potentially more accurate models.
VoxtLM \cite{maiti2023voxtlm} use OPT, which is an open-source LLM \cite{zhang2022opt},  training it on unsupervised speech data with balanced sampling, ensuring that the number of speech tokens is comparable to the number of text tokens to address length differences. Our approach, on the other hand, ensures that the loss is independent of token length for each modality (speech, text), which may enhance training stability across batches.

\textbf{Multimodal processing}:
Efforts in training multimodal models have predominantly focused on the integration of text and vision.
An effective approach to combining text and vision is to leverage pretrained single modals for each modality separately. Subsequently, one can either perform cross-attention between these domains within the embedding space \cite{alayrac2022flamingo} or concatenate the embeddings of the two modalities \cite{liu2024visual}. This methodology has been extended to incorporate audio in \cite{wu2023next}.

While various approaches exist for combining latent spaces across different modalities, some efforts have explored the integration of discrete tokens \cite{team2023gemini, ramesh2021zero, kim2024tmt}. Specifically, discrete tokens have been employed in tasks related to vision and text \cite{ramesh2021zero, ge2023making, du2023role}. What sets our work apart is our emphasis on a tri-modal approach, involving speech, vision, and text. In parallel with our research, \cite{kim2024tmt} has also introduced a tri-modal framework. However, our approach distinguishes itself by harnessing the capabilities of LLM and addressing multilingual aspects, particularly in tasks such as speech-to-speech translation. Another advantageous aspect of our model is its ability to facilitate input as a combination of three modalities: speech, vision, and text. In contrast, in \cite{kim2024tmt}, the input is limited to either speech, vision, or text individually, lacking the capability to accommodate a combination of these modalities.

\section{Discrete Multimodal Language Model}
\label{sec:model}
Our discrete multimodal language model (DMLM) is a  Transformer decoder in the manner of open-source models such as OPT \cite{zhang2022opt}; as such, it can harness knowledge from LLM pretraining on large text corpora.
The model both inputs and outputs discrete tokens across multiple modalities. It can also translate to and from multiple languages. The model can be used for both speech- and non-speech processing tasks.

Let $\mathcal{T}$, $\mathcal{S}$, and $\mathcal{I}$ represent the (mutually disjoint) sets of discrete tokens for the text, speech, and image modalities, respectively, and let $\mathcal{D} =\mathcal{T}\cup\mathcal{S}\cup\mathcal{I}$. These tokens are computed from raw inputs by modality-specific codecs/tokenizers, e.g., Seamless \cite{barrault2023seamlessm4t} for audio, OPT-125 for text, and DALL-E \cite{ramesh2021zero} for images.
Associated with each discrete input token ${\bf x}_i$ is a $d$-dimensional embedding vector, which we can implement  within the first layer of the decoder by extending the existing embedding matrix (of shape $d \times |\mathcal{T}|$) of a pretrained LLM by adding more columns, with a final shape of  $d \times |\mathcal{D}|$. Moreover, the space of possible output tokens is also $\mathcal{D}$. To convert output tokens to speech, we can use a vocoder such as \cite{kong2020hifi}.

We frame each task (ASR, T2S, S2TT, etc.) as a next-token prediction problem. The data is organized as a sequence of tokens ${\bf x}_1, {\bf x}_2, \ldots$, and depending on the specific task at hand, the number and order of modalities may vary. During training the model to translate from a  modality A to modality B, the general  format is:
\noindent {\em Task InputTokens EndA OutputTokens EndB}

\noindent where EndA and EndB represents the end delimiter tokens for the source and target modality, respectively. Note that more advanced uses of the model are also possible (e.g., input both audio and images to perform context-dependent speech recognition), but we leave these to future work.

\textbf{Loss function}:
Naively applying the cross-entropy loss summed over the multimodal sequence of tokens ${\bf x}_1, {\bf x}_2, \ldots$ leads to poor results due to large differences in the number of tokens between domains. For ASR, for example, the number of audio tokens is typically much larger than the corresponding number of text tokens. To account for these differences, we devised a new loss function that normalizes the lengths of tokens within each modality:
\begin{equation*}
\mathcal{L} = -\frac{\lambda_{\mathcal{S}}}{n_\mathcal{S}} \sum_{j \in \mathcal{S}}{p_j \log\hat{p}_j}  -\frac{\lambda_{\mathcal{T}}}{n_\mathcal{T}} \sum_{j \in \mathcal{T}}{p_j \log\hat{p}_j}
  - \frac{\lambda_{\mathcal{I}}}{n_\mathcal{I}} \sum_{j \in \mathcal{I}}{p_j  \log\hat{p}_j}
  \label{equation:eq1}
\end{equation*}
where $\hat{p}$ is the vector of softmax probabilities over the entire output token space ($\mathcal{D}$), and $\lambda_{\mathcal{T}}$, $\lambda_{\mathcal{S}}$, and $\lambda_{I}$ are the weights of text, speech, and image modalities, respectively. Meanwhile, $n_{\mathcal{T}}$, $n_{\mathcal{S}}$, $n_{\mathcal{I}}$ represent the number of text, speech and image tokens. 
This loss function resembles that in DALL-E, albeit with a distinction: In DALL-E \cite{ramesh2021zero}, the loss is weighted between text and vision tokens, and the weights 
$\lambda$ are determined heuristically. In our approach, we utilize hyperparameter search to determine the contribution of each modality to the loss.

\begin{figure}
    \includegraphics[width=\columnwidth]{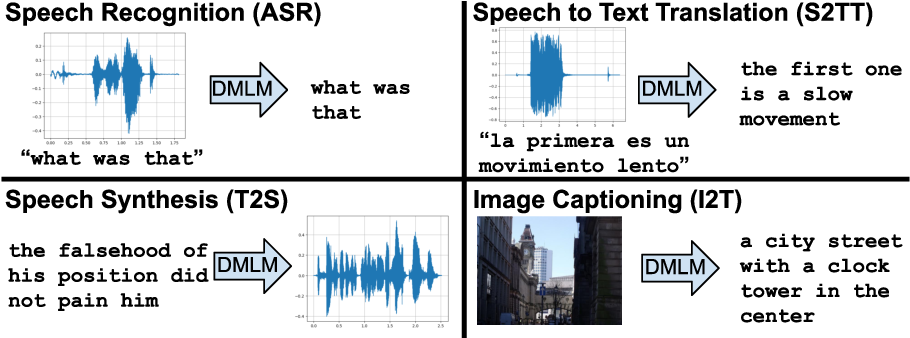}
    \caption{Example inputs/outputs of a trained DMLM.}
    \label{fig:examples}
\end{figure}

\textbf{Example inputs \& outputs}:
Figure \ref{fig:examples} shows example inputs \& outputs of a  DMLM model, trained on the datasets mentioned in Section \ref{sec:experiments}, when applied to different tasks. Note that for these examples, different model weights were optimized for each task; however, in principle a single set of DMLM model weights could be trained to perform all tasks simultaneously.

Also, for the specific task of ASR, we performed supervised training LibriSpeech (960 hours), TED-LIUM (777 hours), SPGI (5000 hours), along with unsupervised on a 3\% subset of Gigaspeech, and thereby produced an ASR model that attains 10.6\% word error rate (WER) on LibriSpeech-other. This is slightly better than the \cite{maiti2023voxtlm} VOXTLM model, which was trained on 4x as much supervised and 160x as much unsupervised data and attains a WER of 11.9\%.

\section{Experiments}
\label{sec:experiments}
Before presenting experiments on (1) the loss function weights $\lambda$, (2) model pretraining, (3) unsupervised training, and (4) tokenization codebooks, we first describe the experimental setup.

{\bf Datasets}:
For speech recognition ({\bf ASR}), we used the following datasets: (1) Librispeech  \cite{panayotov2015librispeech}, encompassing 960 hours (460 hours clean, 500 hours noisy) of speech from public-domain audio books; (2) AMI Meeting \cite{carletta2005ami}, featuring 100 hours of speech from meetings; (3) TED-LIUM ASR corpus \cite{hernandez2018ted} containing a total of 777 hours of recordings of TED talks; 
(4) SPGISpeech \cite{o2021spgispeech}, encompassing 5000 hours of corporate earnings calls along with their associated transcriptions; and (5) Gigaspeech \cite{chen2021gigaspeech}. 
For text-to-speech ({\bf T2S}) tasks, we used LibriTTS-R  \cite{koizumi2023libritts}, containing  585 hours of speech.
For the speech-to-text-translation ({\bf S2TT}) tasks, we used the CoVoST2 dataset \cite{wang2020covost}. CoVoST2 consists of 861 hours of transcribed speech from 21 languages and the corresponding English translation.
Finally, for  image captioning ({\bf I2T})  we used  COCO  \cite{lin2014microsoft}.

{\bf Network, Hyperparameter Settings, \& Codecs}:
We employ the OPT model \cite{zhang2022opt} as our pretrained large language model (LLM) with 125 million parameters. We leverage Hugging Face \cite{wolf2019huggingface} as our training framework. Our experiments run with a batch size of 4, a learning rate set at $5 \times 10^{-5}$, weight decay of $0.0001$, and a patience of 1. Model training is conducted on a single Nvidia $40$ GB or $80$ GB A100 GPU. 

To convert from raw waveform to discrete speech tokens, we use the pretrained Seamless codec \cite{barrault2023seamlessm4t} except in Experiment 4, where we compare to Whisper \cite{radford2023robust}.
To convert from images to discrete tokens and vice-versa, we use DALL-E \cite{ramesh2021zero}.

{\bf Evaluation and Metrics}:
For speech recognition, we utilize the standard WER metric. For text to speech, we transcribe the synthesized speech and use pretrained Whisper large-v2 to transcribe and evaluate the character error rate (CER). For image captioning, we use CIDEr \cite{vedantam2015cider}, BLEU-4 \cite{papineni2002bleu} and METEOR \cite{banerjee2005meteor}. For speech-to-text translation, we use BLEU-4 and sacreBLEU \cite{post-2018-call} 

\subsection{Experiment 1: Tri-Modal Loss Function}

\begin{table}
  \caption{Effect of  $\lambda_\mathcal{S},\lambda_\mathcal{T}$ on ASR WER(\%) of LibriSpeech-clean test set, after training on LibriSpeech (100 hours).}
  \label{tab:hyper_params_search}
  \centering
  \begin{tabular}{ccc}
    \toprule
    \multicolumn{1}{c}{\textbf{Speech weight $\lambda_\mathcal{S}$}} & 
                                         \multicolumn{1}{c}{\textbf{Text weight $\lambda_\mathcal{T}$}} &\multicolumn{1}{c}{\textbf{LS-c}}\\
    \midrule
    0.25 &	0.93  & {\bf 12.9}\\
1.00 & 1.00 & 21.9 \\
0.00 & 1.00 &  15.7 \\
    \bottomrule
  \end{tabular} 
\end{table}
Early during development of DMLM we found that the loss function played a critical role in ensuring the stability of training. In particular, it was necessary to select the  $\lambda$ to account for the varying lengths of tokens depending on modality. Hence, in our first experiment, we assess how $\lambda_\mathcal{S},\lambda_\mathcal{T}$ affect the test accuracy of DMLM when applied to speech recognition. In particular, we train DMLM for ASR on 100 hours of LibriSpeech, and we
compare (1) heuristically setting $\lambda_\mathcal{S}=\lambda_\mathcal{T}$ (similar to Votxtlm \cite{maiti2023voxtlm}); (2) setting $\lambda_\mathcal{S}=0, \lambda_\mathcal{T}=1.00$ (as with Google AudioPaLM \cite{rubenstein2023audiopalm}); and (3) automatic hyperparameter selection using Ax (\texttt{https://ax.dev}).
With Ax, we conducted 25 trials using early stopping on the LibriSpeech-clean development set with WER as the metric to choose the best model.

\textbf{Results}:
Ax selected hyperparameters 
$\lambda_\mathcal{S} =0.25$ and $\lambda_\mathcal{T}=0.93$. We thus used these values in this as well as all other experiments in this paper. Furthermore, as a heuristic choice, we also set $\lambda_\mathcal{I}=0.25$ for image captioning in Experiment 3.

Table~\ref{tab:hyper_params_search} shows the Word Error Rate (WER) on the LibriSpeech-clean test set for different values of $\lambda_\mathcal{S},\lambda_\mathcal{T}$. It is clear that applying a loss to the predictions of both speech and text tokens yields the best WER (12.9\%), outperforming both the setting (used in AudioPalm \cite{rubenstein2023audiopalm}) where only text is modeled (15.7\%) and the setting (used in VoxtLM \cite{maiti2023voxtlm}) with 
balanced weights between speech and text (21.9\%).

\subsection{Experiment 2: Pretraining the LLM}
A potential benefit of a decoder-only model such as DMLM is that it can harness the knowledge from pre-trained LLMs such as OPT. Here, we compare a DMLM 
with random initialization to a model initialized from OPT. In particular, we train DMLM for speech recognition (ASR) on all 960 hours of LibriSpeech, and we evaluated it on the LibriSpeech-clean, LibriSpeech-other, and AMI test sets. 

\textbf{Results}:
The results in Table~\ref{tab:scratch} illustrate a consistent accuracy improvement due to LLM pre-training. 
Notably, the  LLM pre-training makes a large difference for the out-of-domain AMI dataset 
(WER of 94.8\% vs.~658.7\%).

\begin{table}
  \caption{Effect of model pretraining  on ASR WER(\%), comparing random initial model weights to OPT-pretrained weights, followed by training/fine-tuning on LibriSpeech (960 hours).}
  \label{tab:scratch}
  \centering
  \begin{tabular}{ cccc}
    \toprule
    \multicolumn{1}{c}{\textbf{Initialization}} & 
    \multicolumn{1}{c}{\textbf{LS-clean}}&
    \multicolumn{1}{c}{\textbf{LS-other}}&
    \multicolumn{1}{c}{\textbf{AMI}}\\
    \midrule
   Random &	7.1 &	14.3 & 658.7 \\
With LLM &	\textbf{5.6} &	\textbf{12.5} & \textbf{94.8} \\
    \bottomrule
  \end{tabular} 
\end{table}

\subsection{Experiment 3: Mixed-Supervision Training}
In contrast to state-of-the-art speech recognition systems such as Whisper, DMLM  can be trained in an unsupervised manner for next-token prediction, without needing labeled pairs that link across modalities. In this experiment we explore the utility of mixed-supervision training for speech recognition (ASR), speech synthesis (T2S), and image captioning (I2T) tasks.

For {\bf ASR}, we compare 
training only supervised on LibriSpeech (either 100 hours or 960 hours), to training  both supervised on LibriSpeech as well as unsupervised with either speech-only data from the AMI training set, LibriSpeech-500 speech-only data, or Gigaspeech (we used just 3\% of the whole dataset due to computational constraints) text-only data.
For {\bf T2S}, we train either just supervised on LibriTTS-R, or both supervised and unsupervised using Gigaspeech (4\% of the whole dataset). For {\bf I2T}, we compare training just supervised on  COCO, to supervised on COCO combined with LibriSpeech 960 (text-only) data.

\textbf{Results}:
\begin{table}
  \caption{Effect of unsupervised training (plus  supervised training on LibriSpeech-100) on ASR WER(\%) on the LibriSpeech-other test set.}
  \label{tab:unsupervisedLS}
  \centering
  \begin{tabular}{ cc|c}
    \toprule
    \multicolumn{2}{c|}{\textbf{Training set}} & \textbf{Test set} \\
    \multicolumn{1}{c}{\textbf{Supervised  }} & 
    \multicolumn{1}{c|}{\textbf{Unsupervised }} & 
    \multicolumn{1}{c}{\textbf{LS-o}} \\
    \midrule
    LS-100 &	None& 	28.6  \\
     LS-100 &	20\% of LS-500 (Speech only)	& \textbf{25.1} \\
    \bottomrule
  \end{tabular} 
\end{table}
\begin{table}
  \caption{
  Effect of unsupervised training (plus supervised training on LibriSpeech-960) on ASR WER(\%) on the AMI test set.}
  \label{tab:unsupervisedAMI}
  \centering
  \begin{tabular}{ cc|c}
    \toprule
\multicolumn{2}{c|}{\textbf{Training set}} & \textbf{Test set} \\
    \multicolumn{1}{c}{\textbf{Supervised  }} & 
    \multicolumn{1}{c|}{\textbf{Unsupervised }} & 
    \multicolumn{1}{c}{\textbf{AMI}}\\
    \midrule
LS-960 &	None& 94.8 \\
LS-960 &	AMI (Speech only) &  92.6 \\
LS-960 &	3\%  Gigaspeech (Text only)  & \textbf{59.3} \\
    \bottomrule
  \end{tabular} 
\end{table}
Table~\ref{tab:unsupervisedLS} shows that unsupervised training with speech-only data from LibriSpeech-500, in conjunction with supervised training on LibriSpeech-100, improves WER on LibriSpeech-other for  {\bf ASR}.
Similarly, Table \ref{tab:unsupervisedAMI} shows that leveraging speech-only data from AMI, in conjunction with supervised training on LibriSpeech-960, improves WER on the AMI test set. Notably, unsupervised training on a text-only dataset (3\% of Gigaspeech)  decreases the WER even further to 59.3\%.

\begin{table}
  \caption{Effect of unsupervised training (plus supervised training on LibriTTS-R) on T2S CER(\%) on the LibriTTS-R test set}
  \label{tab:TTS}
  \centering
  \begin{tabular}{ cc|c}
    \toprule
\multicolumn{2}{c|}{\textbf{Training set}} & \textbf{Test set} \\
    \multicolumn{1}{c}{\textbf{Supervised  }} & 
    \multicolumn{1}{c|}{\textbf{Unsupervised }} & 
    \multicolumn{1}{c}{\textbf{LibriTTS-R}}\\
    \midrule
LibriTTS-R &	None& 22.9 \\
LibriTTS-R  &	4\%  Gigaspeech (Speech only)  & \textbf{20.2} \\
    \bottomrule
  \end{tabular} 
\end{table}

\begin{table}
\setlength{\tabcolsep}{2pt}
  \caption{Effect of mixed-supervision training on S2TT B@4 (BLEU-4) and sacreBLEU. BLEU and sacreBLEU scores are calculated based on the average performance across 21 languages within the CoVoST2 X$\rightarrow$en test set.}
  \label{tab:covost2}
  \centering
  \begin{tabular}{cc|cc}
    \toprule
\multicolumn{2}{c|}{\textbf{Training set}} & \multicolumn{2}{c}{\textbf{Test set}} \\
    \multicolumn{1}{c}{\textbf{Supervised  }} & 
    \multicolumn{1}{c|}{\textbf{Unsupervised }} & 
    \multicolumn{1}{c}{\textbf{B@4($\uparrow$)}}&
    \multicolumn{1}{c}{\textbf{sacreBLEU ($\uparrow$)}}
    \\
    \midrule
CoVoST2 &	None& 5.71 & 5.75\\
CoVoST2 &	LS-960 (speech)  & 5.74 & 5.76\\
\begin{tabular}[c]{@{}c@{}}CoVoST2 (S2TT)\\+ LS-960 (ASR)\end{tabular}&	None&  \textbf{6.18} & \textbf{6.22}\\
    \bottomrule
  \end{tabular} 
\end{table}

\begin{table}
\setlength{\tabcolsep}{3pt}
  \caption{Effect of unsupervised training (plus supervised training on COCO) on I2T metrics on the COCO test set. B@4: BLEU-4, M: METEOR.}
  \label{tab:image_captioning}
  \centering
\begin{tabular}{ cc|ccc}
    \toprule
\multicolumn{2}{c|}{\textbf{Training set}} & \multicolumn{3}{c}{\textbf{Test set}}  \\
    \multicolumn{1}{c}{\textbf{Supervised  }} & 
    \multicolumn{1}{c|}{\textbf{Unsupervised}} 
        	& \textbf{CIDEr} ($\uparrow$) & \textbf{B@4} ($\uparrow$) & \textbf{M} ($\uparrow$)\\
    \midrule
COCO &	None& 58.1 & 19.5 & 19.2 \\
COCO &	LS-960 (text) & 61.5 &20.6 & 19.9\\
COCO &  SBU (image) & \textbf{62.8} & \textbf{21.1} & \textbf{20.1} \\
    \bottomrule
  \end{tabular} 
\end{table}

{\bf T2S}:
Table~\ref{tab:TTS} shows that, by adding additional 4\% of Gigaspeech for  speech-only unsupervised training, the CER is reduced from 22.9\% to 20.2\%.

{\bf S2TT}:
Table~\ref{tab:covost2} shows a small improvement of speech-to-text translation with unsupervised speech-only data from LibriSpeech-960 (line \#2), as well as a  larger improvement (line \#3) by mixing two forms of supervised training (CoVoST2 for S2TT and LibriSpeech-960 for ASR).

{\bf I2C}: Table~\ref{tab:image_captioning} shows that, by adding the text-only data from LS-960, we improve the CIDEr score from 58.1 to 61.5.  Also, including image-only data from SBU further elevates the CIDEr score to 62.8.
\subsection{Experiment 4: Codebook}
Finally, we investigate the use of an alternative codebook of speech tokens derived from the state-of-the-art Whisper \cite{radford2023robust} model. We use the same approach as \cite{barrault2023seamlessm4t}, namely K-means clustering on the activations from a single layer of the encoder -- we used the final layer of Whisper large-v2. We use the mini-batch k-means algorithm in scikit-learn \cite{minibatchkmeans} with k=10000 and 64 utterances per minibatch. Tokens are then generated by passing audio to the encoder and assigning a token based on the nearest cluster for each timestep. For comparison, we train equivalent k-means models on XLS-R as used in \cite{barrault2023seamlessm4t}. We use Librispeech-960h to train the k-means for both Whisper and XLS-R.
\textbf{Results}:
\begin{table}
  \caption{WER(\%) using Whisper-derived tokens }
  \label{tab:Whisper}
  \centering
  \begin{tabular}{ ccccc }
    \toprule
         \multicolumn{1}{c}{\textbf{Model}} &
    \multicolumn{1}{c}{\textbf{K-means data}} &
    \multicolumn{1}{c}{\textbf{Training set}} &
    \multicolumn{1}{c}{\textbf{LS-c}} &
    \multicolumn{1}{c}{\textbf{LS-o}}\\
    \midrule
Seamless & 	LS-960 & LS-960  & 31.8 & 61.2 \\
Whisper &  LS-960 & LS-960  & \textbf{14.9}& \textbf{23.1} \\
    \bottomrule
  \end{tabular}
\end{table}
We find that using Whisper features as the basis for discrete tokens yields a lower WER on Librispeech test than when the units are trained on XLS-R (Table \ref{tab:Whisper}). This improvement may be attributed to the fact that Whisper was trained with ASR as one of its tasks, thereby learning a more effective representation for this task.  

\section{Discussion and Conclusions}

We introduced DMLM, a decoder-only framework that can harness pre-trained LLMs and can be used for multiple tasks across multiple modalities. DMLM uses discrete speech, vision and text tokens as the inputs and treats all the tasks as a next-token prediction task, thus enabling the framework to train with mixed supervision in a unified manner. We demonstrate the effectiveness of using unsupervised data on various of tasks: speech recognition, speech synthesis, speech to text translation and image captioning. We further investigate the choice of embedding space leading to discrete speech tokens, in particular between Whisper and XLS-R for speech recognition task.

Our work leads to interesting future research directions, e.g., audiovisual ASR through multimodal input streams. 
\newpage
\bibliographystyle{IEEEtran}
\bibliography{mybib}
\end{document}